\documentclass[sigconf]{acmart}
\AtBeginDocument{%
  }
    
\usepackage{subfigure}
\usepackage{amsmath}
\usepackage{fontawesome}

\acmISBN{978-1-4503-XXXX-X/2018/06}

\begin{document}

%%
%% The "title" command has an optional parameter,
%% allowing the author to define a "short title" to be used in page headers.
\title{MoE Enhanced Federated Learning for Spatiotemporal  Prediction}

\author{
Zhehao Dai$^{* 1}$,
Xiao Han$^{* 2 +}$, 
Zhaolin Deng$^{* 3}$,
Zijian Zhang$^{\dagger 4}$,
Xiangyu Zhao$^{\Delta 5}$,
Guojiang Shen$^{* 6}$,
Xiangjie Kong$^{* 7}$
}
\affiliation{
\institution{$^{*}$Zhejiang University of Technology, Zhejiang Key Laboratory of \\ Visual Information Intelligent Processing, Hangzhou}
\institution{$^{\dagger}$Jilin University, Jilin}
\institution{$^{\Delta}$City University of Hong Kong, Hong Kong SAR}
\country{China}
}
% \email{zhehaodai@outlook.com$^{1}$, hahahenha@gmail.com$^{2}$, \{211123120032$^{3}$, gjshen1975$^{6}$\}@zjut.edu.cn, 
% zhangzijian@jlu.edu.cn$^{4}$,
% xianzhao@cityu.edu.hk$^{5}$,
% xjkong@ieee.org$^{7}$
% }
\thanks{$+$ Corresponding author. Email: hahahenha@gmail.com}

%%
%% By default, the full list of authors will be used in the page
%% headers. Often, this list is too long, and will overlap
%% other information printed in the page headers. This command allows
%% the author to define a more concise list
%% of authors' names for this purpose.
\renewcommand{\shortauthors}{Dai et al.}

%%
%% The abstract is a short summary of the work to be presented in the
%% article.
\begin{abstract}
Traffic prediction is fundamental to intelligent transportation systems and urban computing, yet many cities continue to suffer from traffic data scarcity due to limited sensor deployment and uneven urban development. Cross-city knowledge transfer has thus attracted increasing attention, enabling data-rich cities to assist data-scarce ones. However, centralized approaches raise privacy concerns, while existing federated methods struggle with pronounced spatiotemporal heterogeneity across cities. To address these challenges, we propose MoE-FedTP, a personalized federated cross-city spatiotemporal prediction framework based on lightweight Mixture-of-Experts (MoE) networks. MoE-FedTP first employs spatiotemporal neural networks to extract features from both source and target cities, then introduces a set of expert networks derived from different source cities through partial parameter sharing. A gating mechanism dynamically fuses the experts to capture diverse traffic dynamics, achieving fine-grained modeling of urban heterogeneity while preserving privacy. Experiments on four real-world traffic datasets show that MoE-FedTP consistently outperforms state-of-the-art cross-city and federated learning baselines, demonstrating its effectiveness in enhancing prediction accuracy for data-scarce cities.
\end{abstract}

%%
%% The code below is generated by the tool at http://dl.acm.org/ccs.cfm.
%% Please copy and paste the code instead of the example below.
%%
% \begin{CCSXML}
% <ccs2012>
%  <concept>
%   <concept_id>00000000.0000000.0000000</concept_id>
%   <concept_desc>Do Not Use This Code, Generate the Correct Terms for Your Paper</concept_desc>
%   <concept_significance>500</concept_significance>
%  </concept>
%  <concept>
%   <concept_id>00000000.00000000.00000000</concept_id>
%   <concept_desc>Do Not Use This Code, Generate the Correct Terms for Your Paper</concept_desc>
%   <concept_significance>300</concept_significance>
%  </concept>
%  <concept>
%   <concept_id>00000000.00000000.00000000</concept_id>
%   <concept_desc>Do Not Use This Code, Generate the Correct Terms for Your Paper</concept_desc>
%   <concept_significance>100</concept_significance>
%  </concept>
%  <concept>
%   <concept_id>00000000.00000000.00000000</concept_id>
%   <concept_desc>Do Not Use This Code, Generate the Correct Terms for Your Paper</concept_desc>
%   <concept_significance>100</concept_significance>
%  </concept>
% </ccs2012>
% \end{CCSXML}


% \ccsdesc[500]{Computing methodologies~Artificial intelligence}


% \ccsdesc[300]{Do Not Use This Code~Generate the Correct Terms for Your Paper}
% \ccsdesc{Do Not Use This Code~Generate the Correct Terms for Your Paper}
% \ccsdesc[100]{Do Not Use This Code~Generate the Correct Terms for Your Paper}

%%
%% Keywords. The author(s) should pick words that accurately describe
%% the work being presented. Separate the keywords with commas.
%% A "teaser" image appears between the author and affiliation
%% information and the body of the document, and typically spans the
%% page.

% \begin{teaserfigure}
%   \includegraphics[width=\textwidth]{sampleteaser}
%   \caption{Seattle Mariners at Spring Training, 2010.}
%   \Description{Enjoying the baseball game from the third-base
%   seats. Ichiro Suzuki preparing to bat.}
%   \label{fig:teaser}
% \end{teaserfigure}

% \received{20 February 2007}
% \received[revised]{12 March 2009}
% \received[accepted]{5 June 2009}

%%
%% This command processes the author and affiliation and title
%% information and builds the first part of the formatted document.
\maketitle

\section{Introduction}
Traffic prediction plays a crucial role in intelligent transportation systems and urban computing~\cite{liu2023largest, cao2025spatiotemporal, lei2025st,han2020congestion,shen2021attention,han2025garlic}. However, due to insufficient sensor deployment or limited urban development levels, many cities still face the challenge of traffic data scarcity. Consequently, cross-city traffic knowledge transfer~\cite{tang2022domain, lu2022spatio} has gradually emerged as a research hotspot, leveraging knowledge from data-rich cities to assist data-scarce cities in improving prediction performance~\cite{wang2019cross,liu2019traffic}. While existing centralized approaches can achieve certain effectiveness, they pose risks of privacy leakage. Personalized Federated Learning (PFL)~\cite{zhang2024modeling,zhang2025proxy}, as an emerging paradigm, tailors models or methods for each participating client and only shares model parameters rather than raw data for aggregated updates, enabling cross-city knowledge sharing while preserving privacy and providing new possibilities for collaborative prediction~\cite{meng2021cross,zhang2022spatio}.

However, cross-city traffic prediction still faces a prominent challenge—significant spatiotemporal heterogeneity~\cite{ruan2024cross, wang2024time,han2024deep,han2023mitigating}. To address this issue, existing methods have proposed various mitigation strategies: for instance, pFedCTP~\cite{zhang2024personalized} achieves effective transfer through adaptive inter-layer interpolation mechanisms, while STGP~\cite{hu2024prompt} employs a two-stage prompt learning strategy to significantly improve cross-city adaptation performance. Although these methods alleviate model drift, most still rely on unified prediction strategies and struggle to accommodate the inherent differences among different cities in terms of spatiotemporal evolution patterns, functional area distributions, and travel patterns.

To this end, this paper proposes a personalized federated cross-city spatiotemporal prediction framework based on lightweight Mixture-of-Experts networks (MoE-FedTP). Specifically, we first employ spatiotemporal neural networks to extract spatiotemporal features from both source and target cities, then introduce a lightweight Mixture-of-Experts (MoE) structure. The experts in MoE originate from various source cities and are obtained by sharing partial model parameters. Multiple expert networks model different traffic patterns respectively and are dynamically fused by a gating network, thereby achieving fine-grained characterization of urban heterogeneity. This framework preserves privacy advantages while possessing stronger representation and personalization capabilities, particularly enhancing prediction accuracy for data-scarce cities.
Our key contributions are as follows:

\begin{itemize}
\item[$\bullet$] We propose MoE-FedST, a personalized federated spatiotemporal prediction framework with a lightweight MoE structure to address cross-city heterogeneity.
\item[$\bullet$] We design a cross-city parameter sharing mechanism that summarizes spatiotemporal patterns from source cities as expert networks and dynamically transfers them to target cities, enhancing prediction in data-scarce scenarios.
\item[$\bullet$] We conduct extensive experiments on four real-world traffic datasets, demonstrating that MoE-FedST consistently outperforms state-of-the-art baselines in both accuracy and generalization.

\end{itemize}

% \input{draw/frame}

% \begin{figure*}[t]
%     \centering
%     % \includegraphics[width=0.8\linewidth]{draw/frame.pdf}
%     \includegraphics[width=\textwidth]{draw/frame.pdf}
%     \caption{
%     % The framework overview of \name.
%    The framework overview of MoE-FedST.
%     }
%     \label{fig:2}
% % \vspace{-3mm}
% \end{figure*}

\begin{figure*}[htbp]
\centering
\includegraphics[width=\textwidth]{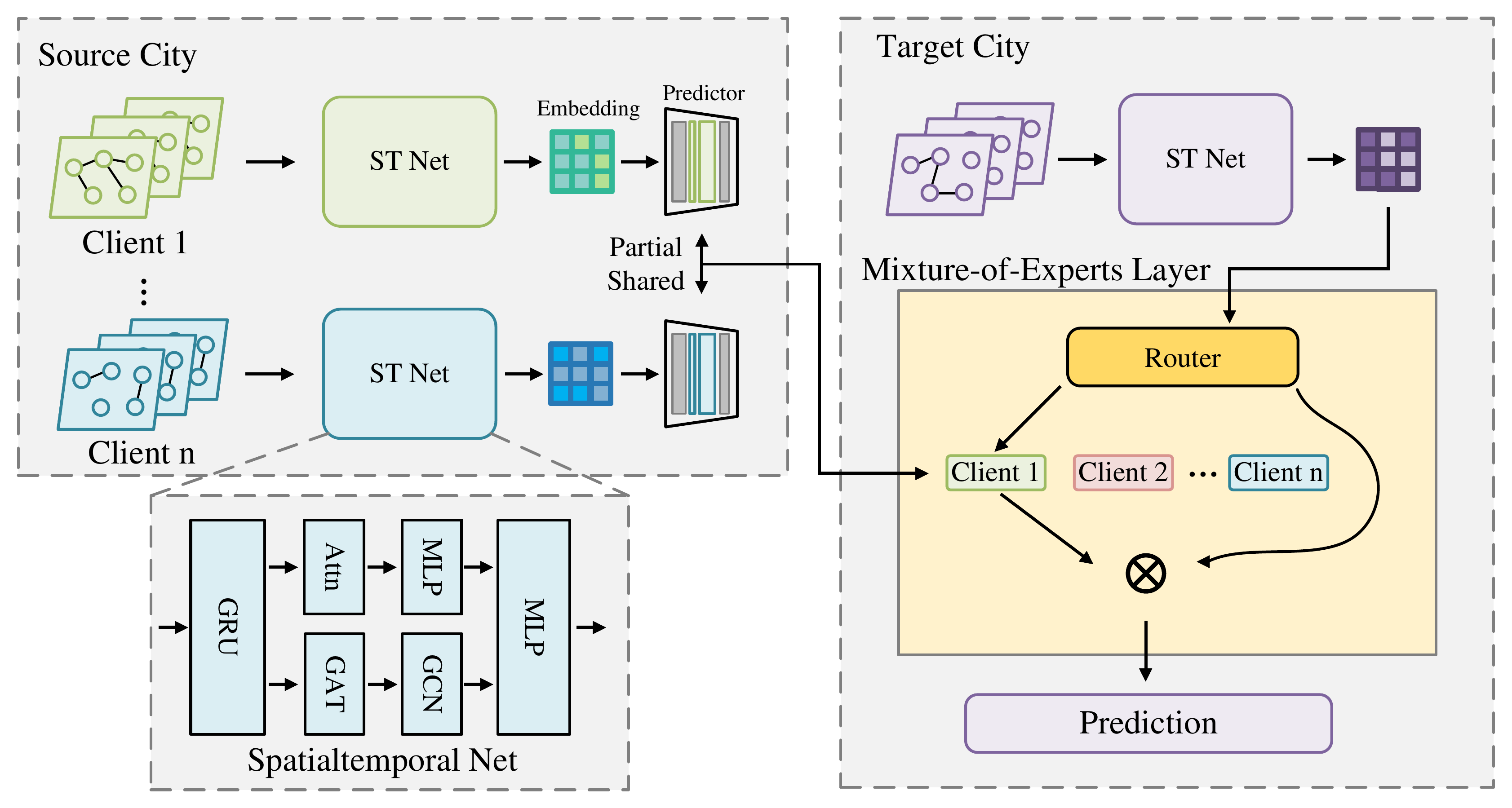}
\caption{The framework overview of MoE-FedsT. It leverages a Spatialtemporal Net to process local spatiotemporal data and employs a MoE model to effectively transfer and integrate learned knowledge from a source city to a target city.}
\Description{}
\label{fig1}
\end{figure*}

\section{ Preliminary}
The sensor distribution in a city can be defined as a graph structure, represented as: $G_{S T}=(\mathcal{V}, \mathcal{E}, \mathcal{A}, X)$, where $\mathcal{V}$ represents the set of nodes, $\mathcal{E}$ denotes the edge set, $\mathcal{A}$ is the adjacency matrix, $\mathcal{A}_{ij}=1$ indicates that a connection exists between node $i$ and node $j$, $X$ is the node feature matrix evolving over time, including information such as traffic volume or speed.

Based on the aforementioned spatio-temporal graph, the forecasting task is formalized as follows: $({X^{1:T}}, G_{st})\stackrel{{ {f(\cdot)}}}{{ \longrightarrow}} {X^{1+T:T+T^{\prime}}}$, where ${X^{1:T}}$ is the node feature sequence of the most recent $T$ time steps, the output $X^{1+T:T+T^{\prime}}$ is the predicted node features for the next $T^{\prime}$ time steps.

Given a model pre-trained on $P$ source domains $G^S = {G^S_1, ..., G^S_P}$ with substantial historical data spanning time period $T_S$, efficiently adapt this model to a new target domain $G^T$ that has limited data over a much shorter period.

\section{Methodology}

In this section, we propose a framework named MoE-FedST for cross-city spatiotemporal forecasting. As illustrated in Figure~\ref{fig1}, the framework comprises two stages. In the source city stage, each city is treated as an independent client with a locally deployed spatiotemporal network, enabling the learning of both shared and city-specific traffic patterns. In the target city stage, partial parameters from source-city predictors are shared as experts in an MoE model. The target city dynamically activates only the top-k relevant experts, ensuring lightweight computation while alleviating data scarcity and spatiotemporal heterogeneity.

\subsection{Local Learning in Source Cities}
To effectively capture spatiotemporal patterns in urban traffic data, we employ a spatiotemporal neural network during local training. As illustrated in the lower-left part of Figure 1, the network is composed of multiple components, including Graph
Convolutional Network (GCN)~\cite{kipf2017semi}, Graph Attention Network (GAT)~\cite{petar2018graph}, and Attention modules, which jointly model the spatial dependencies and temporal dynamics of traffic data. This design enables the source cities to learn more representative features and achieve stronger predictive performance. Mathematically, for an input spatiotemporal feature vector, the output of the spatiotemporal network is computed as:
\begin{equation}
\begin{aligned}
\mathbf{H}_{\textit{temporal}}&=\text{MLP}\left(\text{Att}_{\textit{temp}} \left(\mathbf{H}_{\textit{gru}}\right)\right), \\
\mathbf{\bar{H}}&=\mathrm{Pooling}_{time}(\mathbf{H}_{gru}),\\
\mathbf{H}_{spatial}&=\mathrm{GCN}\left(\mathrm{GAT}(\mathbf{\bar{H}})\right),\\
% \mathbf{H}_{spatial}=\text{GCN(GAT(H}^{(t)}_{gru}))，
\end{aligned}
\end{equation}
where $\mathbf{H}_{\mathrm{gru}}\in\mathbb{R}^{T\times N\times F}$ is the output tensor of the Gated Recurrent Unit (GRU)~\cite{chung2014empirical} network. Here, $T$ is time steps, $N$ represents the number of nodes, and $F$ denotes the feature dimensionality. $\bar{\mathbf{H}}\in\mathbb{R}^{N\times F}$ is the node-level summary feature matrix obtained by applying time-dimensional pooling to the GRU output.

Finally, fuse the temporal and spatial embeddings to obtain the final output embedding:
\begin{equation}
\mathbf{H}_{st} = \mathrm{MLP}(\mathbf{H}_{\textit{temporal}},\mathbf{H}_{spatial}).
\end{equation}
We have designed and developed a comprehensive spatiotemporal predictor that effectively integrates temporal and spatial feature information to achieve accurate traffic flow forecasting for the next $T$ time steps. Through the construction of an MLP-based prediction architecture, this system can deeply fuse embedding representations from both temporal and spatial dimensions, ultimately generating high-precision traffic flow prediction results. $Y'=MLP(H_{st})$, where $Y'$ is the prediction result.

To optimize the model parameters during training, we adopt the Mean Squared Error (MSE) as the loss function to evaluate prediction accuracy. Let $Y\in\mathbb{R}^{N\times T'\times D}$ denote the ground truth traffic data, where $N$ represents the number of sensors, $T'$ indicates the prediction horizon, and $D$ is the feature dimension. The loss function for the traffic prediction model is formulated as: 
\begin{equation}
\mathcal{L}_{pred}=\frac{1}{T^{\prime}}\sum_{t=1}^{T^{\prime}}(Y^t-{Y'}^t)^2.
\end{equation}

\subsection{Personalized Federated Learning in Target Cities}
In the target city adaptation phase, we design a novel personalized federated adaptation mechanism based on MoE architecture. Specifically, we extract critical parameters from the pre-trained prediction heads of multiple source cities, sharing and retaining these parameters as expert networks, where each expert network encodes the unique spatiotemporal traffic patterns and domain knowledge of its respective source city. The gating network computes the importance weights for each expert as:
\begin{equation}
g_i=\frac{\exp(h_i(H^t_{tg}))}{\sum_j\exp(h_j(H^t_{tg}))},\quad i=1,\ldots,M,
\end{equation}
where $M$ is the number of experts and $h_i(\cdot)$ is the gating function. To ensure efficiency, we only activate the top-k experts with the highest gating scores. Let $\mathcal{K}_t\subseteq\{1,\ldots,M\}$ denote the selected experts, then the prediction is obtained as: ${Y'}^t_{tg}=\sum_{i\in\mathcal{K}_t}g_iE_i(X^t_{tg})$. The target city minimizes its own prediction loss:
\begin{equation}
\mathcal{L}_{\mathrm{target}}=\frac{1}{|\mathcal{D}_q|}\sum_{(X^t_{tg},Y^t_{tg})\in\mathcal{D}_q}\mathcal{L}_{pred}({Y'}^t_{tg},Y^t_{tg}),
\end{equation}
where ${D}_q$ indicates the size of the target city dataset.

This selective activation mechanism not only effectively reduces computational complexity and resource consumption, but more importantly, ensures that the model can intelligently leverage expert knowledge most suitable for the current target city's traffic characteristics. Consequently, even under data-scarce constraints, the target city's traffic prediction accuracy can be significantly enhanced while preserving data privacy and effectively mitigating the challenges posed by cross-city spatiotemporal heterogeneity.

\section{Experiments}
\subsection{Experimental Setup}
\noindent \textbf{Datasets.}
\quad
\begin{table}[t]
\centering
% \footnotesize
\small
\caption{Statistics of datasets.}
\begin{tabular}{lcccc}
\toprule
Dataset & \#Sensor & \#Edge & Time Span & Time Interval \\
\midrule
PEMS-BAY & 325 & 2,694 & 52,116 & 5 min \\
METR-LA & 207 & 1,722 & 34,272 & 5 min \\
DiDi-Chengdu & 524 & 1,120 & 17,280 & 10 min \\
DiDiot-Shenzhen & 627 & 4,845 & 17,280 & 10 min \\
\bottomrule
\end{tabular}

\label{tab:datasets}
\end{table}
\begin{table*}[t]
\centering
\small
\caption{Comparative performance of models across datasets. The optimal results are highlighted for clarity.}
\begin{tabular}{lcccccccc}
\toprule
 & \multicolumn{2}{c}{PEMS-BAY} & \multicolumn{2}{c}{METR-LA} & \multicolumn{2}{c}{DiDi-Chengdu} & \multicolumn{2}{c}{DiDi-Shenzhen} \\
\multicolumn{1}{l}{Method} & \multicolumn{2}{c}{5 min/15 min/30 min} & \multicolumn{2}{c}{5 min/15 min/30 min} & \multicolumn{2}{c}{10 min/30 min/60 min} & \multicolumn{2}{c}{10 min/30 min/60 min} \\
 & MAE & RMSE & MAE & RMSE & MAE & RMSE & MAE & RMSE \\ \midrule
STGCN & 2.64/2.77/3.01 & 4.03/4.44/5.13 & 4.26/4.42/4.86 & 6.32/6.67/7.35 & 2.93/2.91/3.09 & 4.25/4.25/4.49 & 2.74/2.74/2.87 & 3.86/3.88/4.09 \\
ST-GFSL & 1.16/1.64/2.17 & 1.98/3.26/4.64 & 2.46/3.11/3.82 & 4.14/5.61/6.93 & 2.19/2.64/\textbf{3.00} & 3.15/\textbf{3.84}/\textbf{4.34} & 1.95/2.36/2.72 & 2.81/3.52/4.13 \\
pFedCTP & \textbf{1.08}/1.60/2.20 & \textbf{1.83}/3.21/4.58 & 2.33/3.01/3.81 & 4.08/5.68/7.07 & \textbf{2.14}/2.66/\textbf{3.00} & \textbf{3.10}/3.90/4.39 & 1.89/2.34/2.70 & 2.75/3.52/4.12 \\  \midrule
MoE-FedST & 1.20/\textbf{1.57}/\textbf{2.00} & 2.10/\textbf{3.20}/\textbf{4.45} & \textbf{2.23}/\textbf{2.88}/\textbf{3.47} & \textbf{3.94}/\textbf{5.43}/\textbf{6.52} & 2.19/\textbf{2.63}/3.08 & 3.18/4.05/4.35 & \textbf{1.88}/\textbf{2.30}/\textbf{2.59} & \textbf{2.70}/\textbf{3.31}/\textbf{3.73} \\
\bottomrule
\end{tabular}
\label{tab:main}
\end{table*}
We comprehensively evaluate our proposed method on real-world traffic flow datasets from four major metropolitan areas, including PEMS-BAY, METR-LA~\cite{li2018diffusion}, DiDi-Chengdu, and DiDi-Shenzhen. The detailed statistics of these datasets are presented in Table~\ref{tab:datasets}. In our experimental design, we adopt a rotation validation strategy: three of the datasets serve as data-rich source cities for pre-training, while the remaining fourth dataset provides only three days of traffic data to simulate a data-scarce target city scenario.

\noindent \textbf{Metrics.}
\quad
We employ Mean Absolute Error (MAE) and Root Mean Square Error (RMSE) as the primary evaluation metrics to assess the performance of our model.

\noindent \textbf{Models.}
\quad
We compare our MoE-FedTP against several representative baseline methods: STGCN~\cite{yu2018spatio} as spatiotemporal graph neural network baselines that capture traffic spatiotemporal dependencies; ST-GFSL~\cite{hu2024prompt} as a few-shot learning approach for cross-city traffic prediction; and pFedCTP~\cite{zhang2024personalized} as a PFL method designed for cross-city scenarios with adaptive interpolation mechanisms. These baselines represent state-of-the-art approaches across different paradigms, including spatiotemporal modeling and federated learning.

\noindent \textbf{Evaluation Settings.}
\quad
We establish uniform experimental settings across all methods for fair evaluation. The temporal configuration includes history step $T = 12$ and prediction step $T'=6$, with evaluation on 1, 3, and 6 step ahead predictions for short, medium, and long term forecasting. Key hyperparameters: batch size = 32, learning rate = 0.01, hidden dimensions = 64. Baseline models are trained for 100 epochs. Our MoE-FedTP employs $M=3$ experts with top-$k = 2$ activation.

\subsection{Performance Evaluation}

We report the comparative analysis between our model and the baselines in Table~\ref{tab:main}. The results demonstrate that MoE-FedST achieves competitive performance across all four datasets, particularly on METR-LA and DiDi-Shenzhen. Specifically, because of the introduction of the MoE network, our model has demonstrated outstanding performance in both long-term and short-term traffic prediction tasks. Specifically, on the METR-LA dataset, the MAE and RMSE metrics improved by 6.23\% and 5.20\%, respectively; while on the DiDi-Shenzhen dataset, an average performance gain of 4.68\% was achieved. These results further validate the strong generalization capability of our method, making it effectively applicable to different urban traffic scenarios.

\subsection{Ablation Study}
\begin{figure}[tbp]
\centering

\setcounter{subfigure}{0}
\subfigure[PEMS-BAY]{
\includegraphics[width=0.47\linewidth]{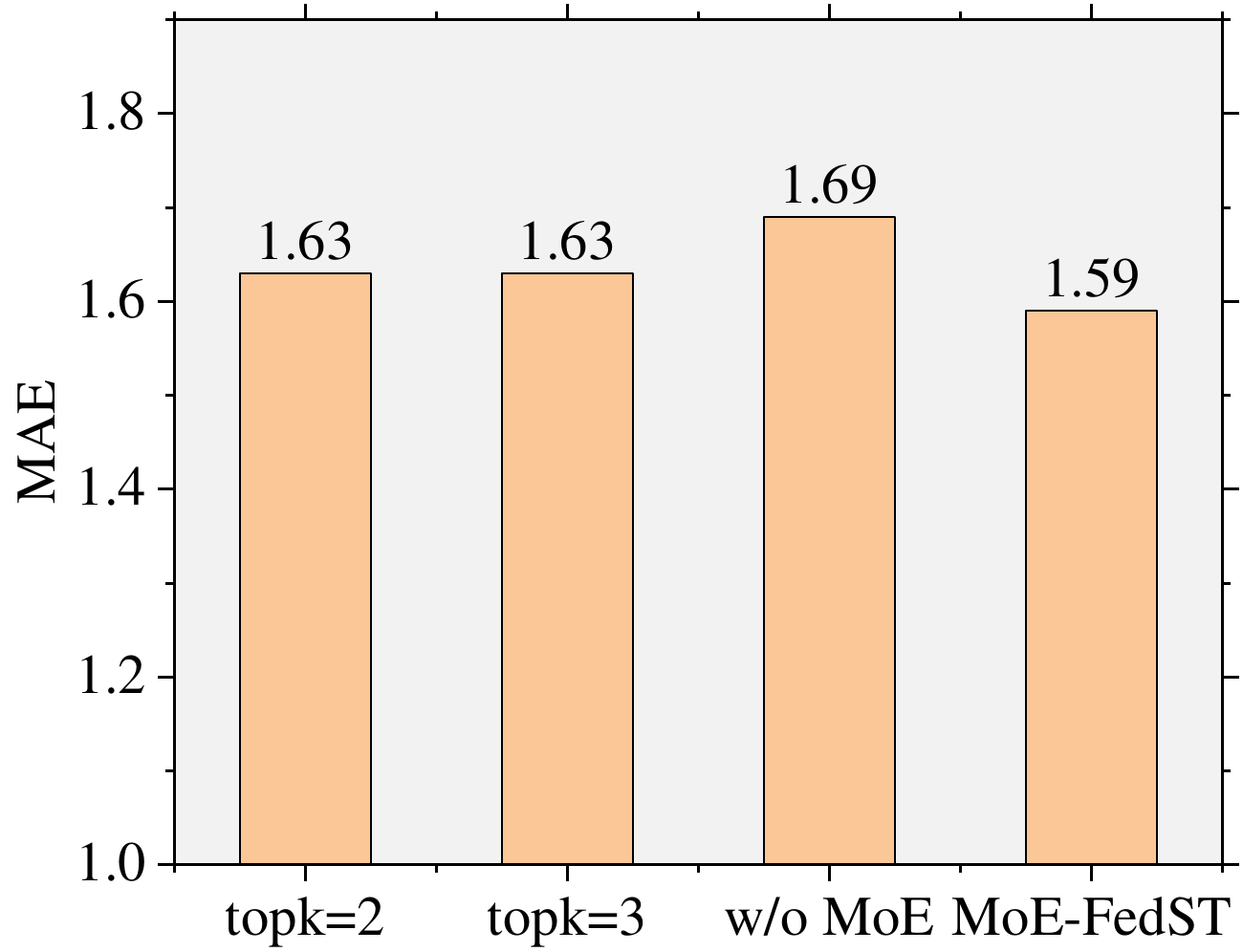}
\label{fig:4a}
}
\subfigure[METR-LA]{
\includegraphics[width=0.47\linewidth]{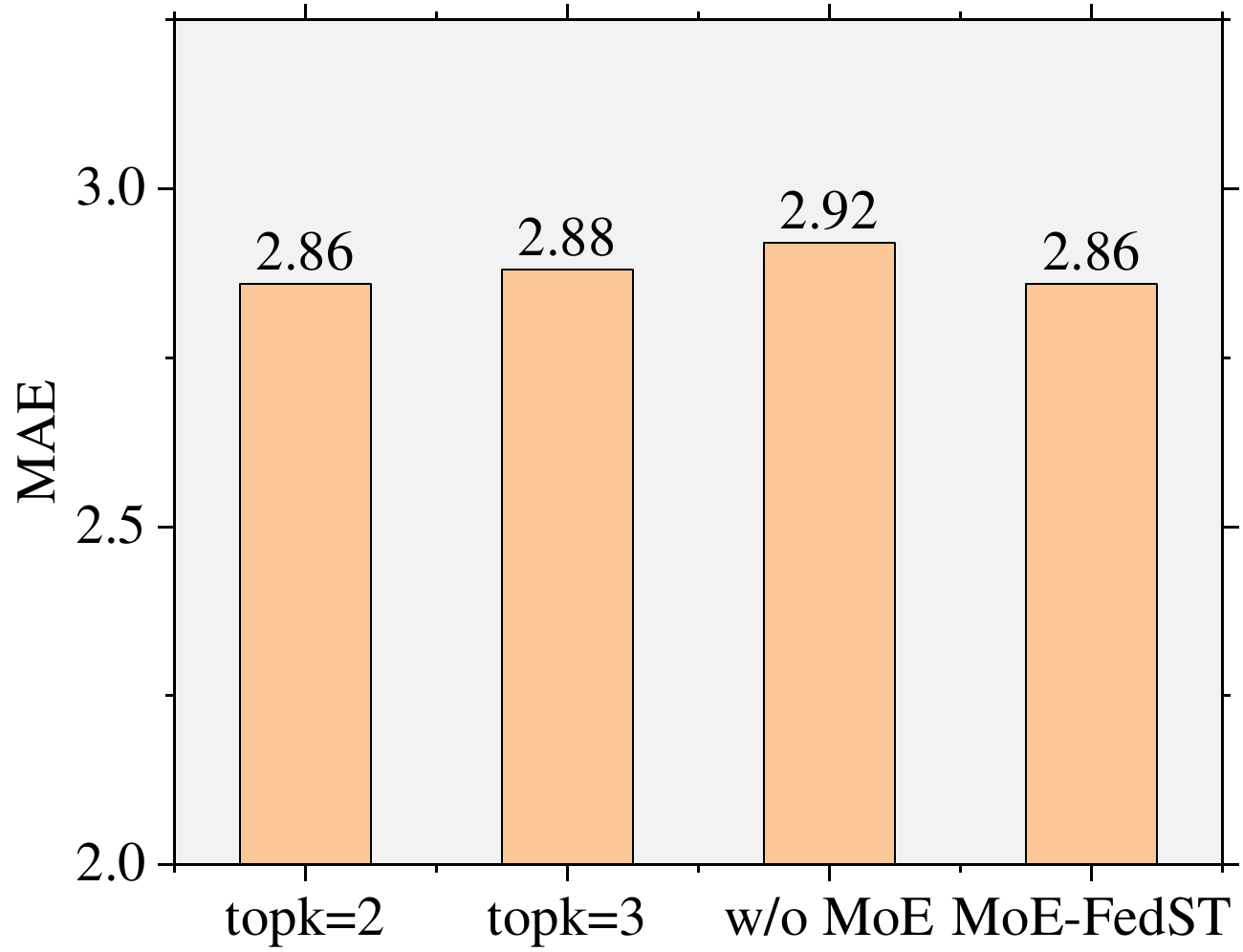}
\label{fig:4b}
}
\subfigure[Didi-Chengdu]{
\includegraphics[width=0.47\linewidth]{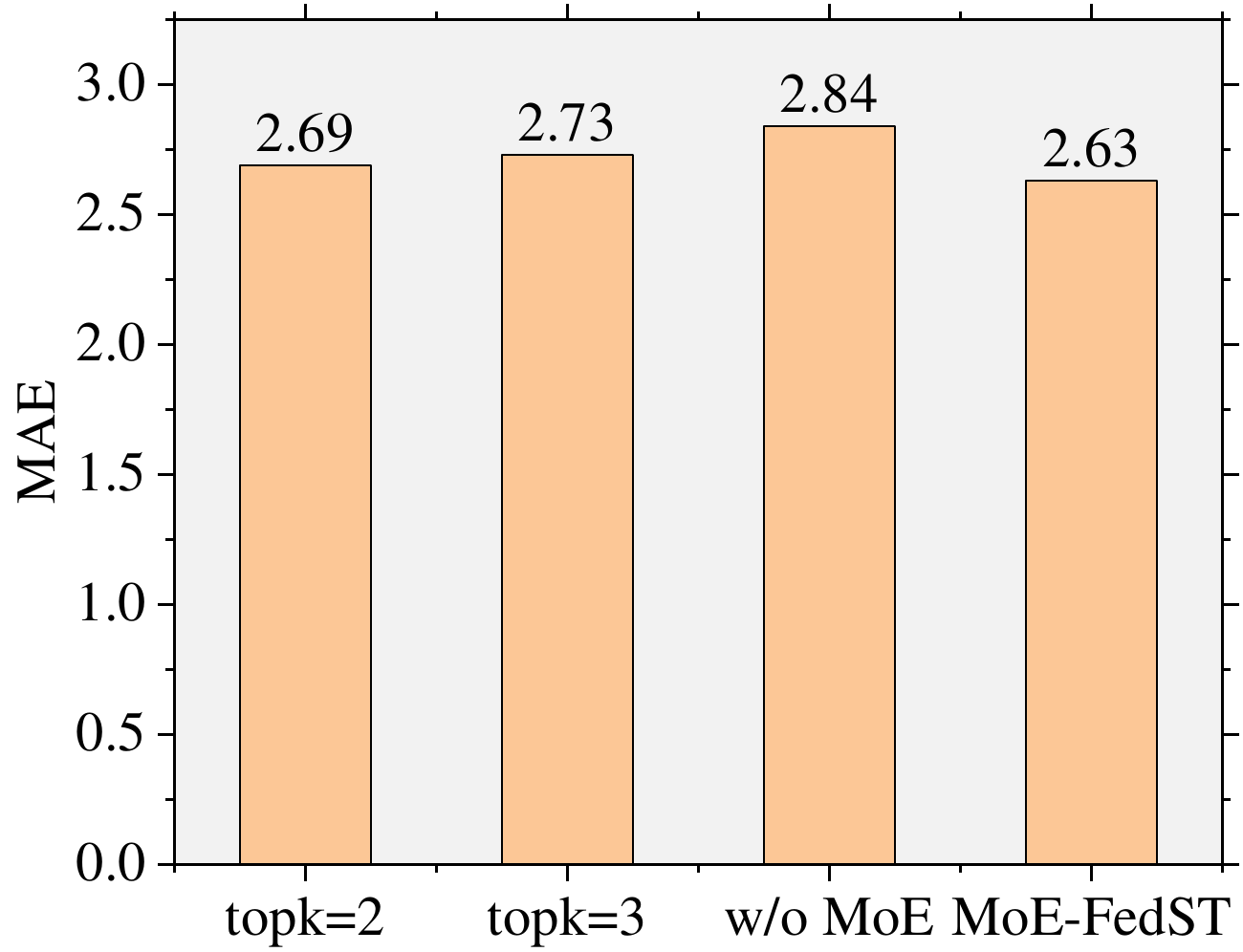}
\label{fig:4c}
}
\subfigure[Didi-Shenzhen]{
\includegraphics[width=0.47\linewidth]{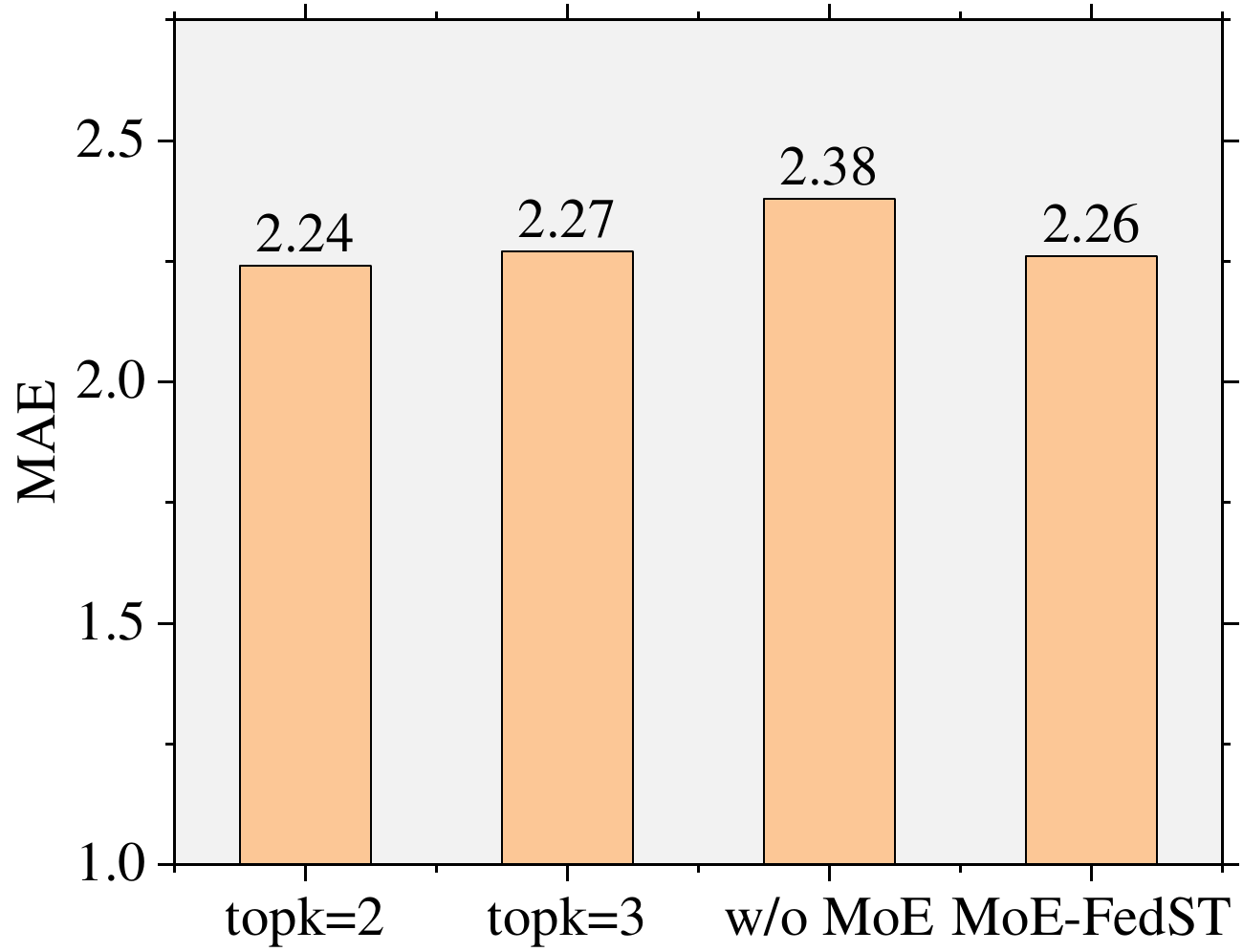}
\label{fig:4d}
}
\caption{Ablation study of MoE-FedST.}
\label{fig:4}
% \vspace{-3mm}
\end{figure}
To evaluate the effectiveness of our model components, we conducted relevant ablation studies by varying the number of experts $k$ selected in the top-$k$ mechanism, removing the expert model, and the spatiotemporal net. As illustrated by the experimental results across the four datasets, we can conclude that the expert network is essential for enhancing the overall performance of the model. 
Specifically, as shown in Figure~\ref{fig:4c} and Figure~\ref{fig:4d}, on the DiDi-Chengdu and DiDi-Shenzhen datasets, the MAE performance degrades by 7.4\% and 5.0\% respectively when the MoE module is removed. Furthermore, in the analysis of the k parameter in top-k expert selection, both DiDi-Chengdu and DiDi-Shenzhen datasets achieve optimal performance when k=1, indicating that for cities with distinctive traffic patterns, selecting the most relevant single expert can better capture their specific spatiotemporal patterns, while activating too many experts may introduce noise and dilute the contribution of the best expert.
In contrast, as shown in Figure~\ref{fig:4a} and Figure~\ref{fig:4b}, the PEMS-BAY and METR-LA datasets perform best when k=2, suggesting that these cities have more complex and diverse traffic patterns that require combining knowledge from multiple experts to achieve optimal prediction performance.

\section{Future Work}
% Although MoE-FedTP shows strong performance in cross-city spatiotemporal prediction, several limitations remain. First, the framework relies on partial parameter sharing across source cities to capture diverse traffic dynamics, which may be insufficient when the target city exhibits extreme heterogeneity or when source cities are limited. Second, while the lightweight MoE design reduces computational overhead, the fixed top-k expert selection strategy may not always adapt optimally to varying traffic scenarios.

% For future work, we plan to explore more adaptive expert selection strategies, such as reinforcement learning-based or uncertainty-aware gating, to further improve personalization. Extending MoE-FedTP to multi-modal urban data and heterogeneous sensor networks is another promising direction.

% Despite strong performance, MoE-FedTP has limitations. The partial parameter sharing may be insufficient for cities with extreme heterogeneity or when source cities are limited. Additionally, the fixed top-k expert selection strategy may not optimally adapt to varying traffic scenarios.
% Future work will explore adaptive expert selection strategies, such as reinforcement learning-based or uncertainty-aware gating, and extend the framework to multi-modal urban data and heterogeneous sensor networks.

Despite its strong performance, MoE-FedTP has notable limitations. The partial parameter sharing mechanism may prove insufficient for cities with extreme heterogeneity in traffic patterns or when the number of source cities is limited, potentially reducing the model’s ability to generalize effectively across diverse environments. Additionally, the fixed top-k expert selection strategy may not optimally adapt to dynamic and varying traffic scenarios. Future work will address these limitations by exploring more adaptive expert selection strategies, such as reinforcement learning-based or uncertainty-aware gating approaches, to enhance the model's flexibility in responding to changing traffic conditions. Furthermore, the framework will be extended to incorporate multi-modal urban data, integrating various transportation modes and heterogeneous sensor networks, to improve its accuracy, scalability, and robustness across a wider range of urban environments.

\section{Conclusion}
% In this paper, we presented MoE-FedTP, a cross-city personalized federated spatiotemporal prediction framework built upon lightweight MoE networks. The framework addresses two key challenges in traffic prediction: the pronounced spatiotemporal heterogeneity across cities and the scarcity of data in many urban regions. Specifically, MoE-FedTP leverages spatiotemporal neural networks to extract informative features from source cities, introduces expert networks through partial parameter sharing, and employs a gating mechanism in target cities to dynamically select and integrate the most relevant experts. This design enables fine-grained modeling of urban heterogeneity while ensuring privacy preservation.
% Extensive experiments on four real-world traffic datasets demonstrate that MoE-FedTP consistently outperforms state-of-the-art cross-city and federated learning baselines in both prediction accuracy and generalization capability. Furthermore, ablation studies confirm the effectiveness of the MoE structure and expert selection mechanism in enhancing model performance under data-scarce conditions.
% Overall, MoE-FedTP provides an efficient and privacy-aware solution for cross-city traffic prediction, offering a promising direction for federated transfer learning in broader urban computing and spatiotemporal forecasting applications.

In this paper, we propose MoE-FedTP, a cross-city personalized federated spatiotemporal prediction framework based on lightweight MoE networks. The framework tackles two major challenges in traffic prediction: pronounced spatiotemporal heterogeneity across cities and data scarcity in many urban regions. MoE-FedTP employs spatiotemporal neural networks for feature extraction, constructs expert networks via partial parameter sharing, and applies a gating mechanism to dynamically select and fuse the most relevant experts, enabling fine-grained modeling of urban heterogeneity while preserving privacy.
Extensive experiments on four real-world traffic datasets show that MoE-FedTP consistently outperforms state-of-the-art cross-city and federated learning baselines in both accuracy and generalization. Ablation studies further confirm the effectiveness of the MoE structure and expert selection strategy under data-scarce conditions. Overall, MoE-FedTP provides an efficient and privacy-preserving solution for cross-city traffic prediction and offers promising directions for federated transfer learning in broader urban computing and spatiotemporal forecasting tasks.

\appendix

\end{document}